\documentclass[10pt, conference]{IEEEtran}

\usepackage{amsmath,amssymb,amsfonts}
\usepackage{graphicx}
\usepackage{pifont}

\usepackage{amsthm}
\usepackage{xtab,xcolor}
\usepackage[figuresright]{rotating}

\usepackage{graphicx}
\graphicspath{{/}{fig/}}

\usepackage{array}
\usepackage{textcomp}
\usepackage{xcolor}
\usepackage{multirow}
\usepackage{booktabs}

\usepackage{mathtools}
\usepackage{breqn}
\usepackage{float}
\usepackage{gensymb}

\usepackage{pgfplots}
\pgfplotsset{compat=1.7}
\usepgfplotslibrary{groupplots}

\usepackage{caption}
\usepackage{subcaption}

\usepackage{multirow,tabularx}
\usepackage{hyperref}
\usepackage{flushend}
\usepackage{algorithmic}
\usepackage[vlined, ruled, shortend]{algorithm2e}

\usepackage{longtable}

\newlength\figureheight
\newlength\figurewidth
\SetAlCapNameFnt{\footnotesize}
\SetAlCapFnt{\footnotesize}

\title{
    Vision-based GNSS-Free Localization for \\ UAVs in the Wild
}

\author{
    \IEEEauthorblockN{
        \vspace{1em}
        Marius-Mihail Gurgu\IEEEauthorrefmark{2},
        Jorge Peña Queralta\IEEEauthorrefmark{2},
        Tomi Westerlund\IEEEauthorrefmark{2}
    }
    \IEEEauthorblockA{
        \normalsize
        \IEEEauthorrefmark{2}\href{https://tiers.utu.fi}{Turku Intelligent Embedded and Robotic Systems (TIERS) Lab}\\ University of Turku, Finland.\\
        Emails: \textsuperscript{1}\{mmgurg, jopequ, tovewe\}@utu.fi\\[+6pt]
    }
}

\begin{document}

\maketitle
\thispagestyle{empty}
\pagestyle{empty}



\begin{abstract}%
    \label{sec:abstract}%
    Considering the accelerated development of Unmanned Aerial Vehicles (UAVs) applications in both industrial and research scenarios, there is an increasing need for localizing these aerial systems in non-urban environments, using GNSS-Free, vision-based methods. Our paper proposes a vision-based localization algorithm that utilizes deep features to compute geographical coordinates of a UAV flying in the wild. The method is based on matching salient features of RGB photographs captured by the drone camera and sections of a pre-built map consisting of georeferenced open-source satellite images. Experimental results prove that vision-based localization has comparable accuracy with traditional GNSS-based methods, which serve as ground truth. Compared to state-of-the-art Visual Odometry (VO) approaches, our solution is designed for long-distance, high-altitude UAV flights. Code and datasets are available at \url{https://github.com/TIERS/wildnav}.

\end{abstract}

\begin{IEEEkeywords}

    UAV; MAV;  GNSS-Free; GNSS-denied; Vision-based localization; Photogrammetry; Computer vision; Perception-based localization; Visual Odometry

\end{IEEEkeywords}
\IEEEpeerreviewmaketitle


\section{Introduction}\label{sec:introduction}

Unmanned Aerial Vehicles (UAVs) are currently used in a wide range of scenarios and commercial applications. They already have reliable localization methods based on the Global Navigation Satellite System (GNSS)~\cite{alsalam2017autonomous}, Visual-Inertial Simultaneous Localization and Mapping (VI-SLAM)~\cite{qin2018vins}, or Ultra-wideband (UWB) systems~\cite{xianjia2021cooperative}. The first method is usually used  in outdoor environments, where it is relatively simple to use a GNSS sensor to obtain accurate positioning data~\cite{tomavstik2019uav}. The second  and third methods prove their utility in environments where GNSS signal is unreliable, for example inside buildings, where localization systems such as UWB~\cite{queralta2020uwb, queralta2022viouwb} or VI-SLAM are preferred~\cite{tordesillas2021panther}. {\let\thefootnote\relax\footnote{{------------------------------------------------------------------------------------------This research work is supported by the Academy of Finland's AutoSOS and AeroPolis projects (Grant No. 328755 and 348480) and by the Finnish Ministry of Defence's Scientific Advisory Board for Defence (MATINE) project WildNav.}}}

Autonomy of aerial drones and their hardware capability have significantly improved in the last years~\cite{nguyen2021viral}, enabling them to safely fly autonomously over relatively long distances, beyond visual line of sight (BVLOS). However, less attention had been given to providing accurate positioning data during long distance UAV flights without using GNSS, which would lead to more resilient solutions. This is an essential matter since UAVs are mission-critical systems in need of a failsafe mechanism that can be automatically engaged when GNSS signal becomes unavailable due to various reasons, such as jamming or spoofing~\cite{7445815}.

At the same time, image processing capabilities using deep neural networks have been constantly improving in the last decade, enabling even real-time recognition of objects and segmentation. This observation also holds for the processing of satellite image data, where artificial intelligence (AI) models are trained to differentiate between important objects such as buildings and rivers~\cite{shermeyer2020spacenet}. 

Building on top of available open source technology and algorithms~\cite{sarlin2020superglue, yol2014vision}, this paper aims to provide a reliable positioning mechanism for UAVs by only using  RGB photographs provided by a camera mounted on the drone and open-source satellite images of the flight area. The method is based on deep neural image segmentation  that enables the localization of the UAV by extracting recognizable features of buildings, roads, rivers and forest edges. These features are further matched to static satellite images using a robust method based on a graph neural network model~\cite{sarlin2020superglue}.

\begin{figure}
    \centering
    \includegraphics[width=0.49\textwidth]{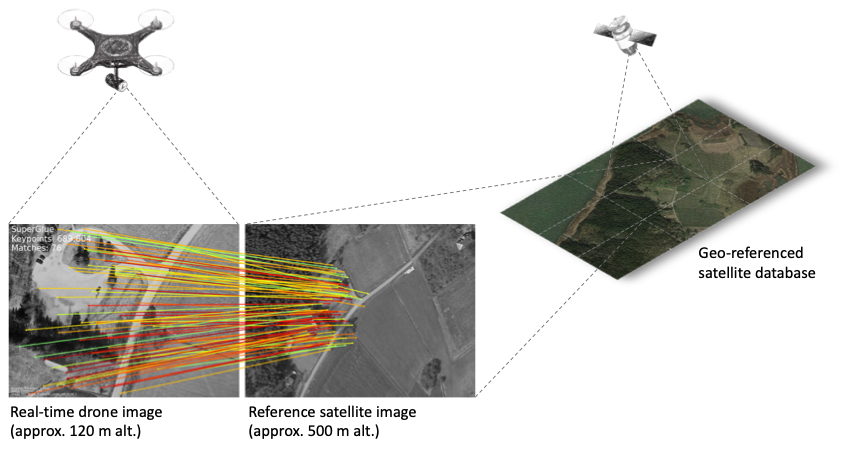}
    \caption{Conceptual illustration of the proposed approach. Drone images are compared to a set of reference satellite images and a match is found based on deep features from Superglue~\cite{sarlin2020superglue}.}
    \label{fig:concept}
\end{figure}

Fig.~\ref{fig:concept} shows the core idea of the proposed visual-based localization method, where an UAV flying at a high altitude – provided that the ground surface below it has enough landmarks with salient features – is able to match the camera stream with onboard, georeferenced, open-source satellite images.

The paper is organized in 6 sections as follows. Section~\ref{sec:related_work} provides a brief overview of related works on which the current paper is based. Section~\ref{sec:problem_to_solve} explains the motivation and applicability of the implemented vision-based algorithm, while Section~\ref{sec:methodology} presents the approach used for determining the absolute geographical coordinates of a UAV. Section~\ref{sec:results} discusses the experimental localization results. Finally, Section~\ref{sec:conclusion} concludes the work and outlines future research directions.


\section{Background} \label{sec:related_work}

Previous studies directly related to our paper include semantic segmentation based path planning~\cite{bartolomei2020perception}, localization using open source Google Earth (GE) aerial images~\cite{patel2020visual} and pose estimation with neural networks trained with georeferenced satellite photographs~\cite{shetty2019uav}. In addition to these, an open source image segmentation model originally used for building virtual worlds~\cite{guerin2021satellite} can prove its utility in providing useful input both for localization and navigation purposes, as proposed in~\cite{bartolomei2020perception}.

Additionally, a graph neural network that provides feature computation and matching for outdoor images, SuperGlue~\cite{sarlin2020superglue}, became part of the implementation in our proposed localization algorithm due to its remarkable performance of matching features in photographs which significantly differ in perspective and lighting conditions. The model proved its efficient application not only for matching features, but also for the perspective transformations (namely homography) used in computing geographical coordinates when running the implemented vision-based localization algorithm. Before selecting Superglue as a feature matcher between drone camera photographs and satellite images, template matching~\cite{briechle2001template} and SIFT features~\cite{lowe2004distinctive} were also taken into consideration. The latter two options were dropped due to their relatively high computation time and inaccuracy, compared to~\cite{sarlin2020superglue}.

Another remarkable  paper is~\cite{yol2014vision}, which uses an advanced template-based matching algorithm to compute the pose estimation of a UAV using only images. Because of insufficient computation resources at the time,  no implementation is provided on the onboard computer of a drone. Another major difference, compared to our approach, is that the drone navigates specifically urban environments, where computing and matching visual features is much less of a challenge, compared to natural, \textit{in the wild}, flying areas.

Currently, the state-of-the-art of navigation approaches in \textit{a priori} unknown environments are Simultaneous Localization and Mapping (SLAM) algorithms, which have arguably reached a high-degree of robustness and reliability in the past decade~\cite{macario2022comprehensive}. An important building block of SLAM algorithms are VO methods~\cite{delmerico2018benchmark}, which allow UAVs to accurately determine their position while navigating new environments. However, one major limiting factor of these approaches is that they make the assumption the UAV is flying at a low enough altitude that an RGB monocular or stereo camera can easily track position shifts in detected features from frame to frame. Our work focuses on high-altitude flight (120 meters), a situation where a different approach for localization is needed, as presented in Section \ref{sec:problem_to_solve}.


\section{Providing GNSS-Free vision-based localization}
\label{sec:problem_to_solve}

Our main goal is developing a localization algorithm that does not rely on GNSS for long-distance flights, but only on a monocular wide-angle camera. This kind of approach proves its utility in situations where GNSS signal cannot be reliably used and as a failsafe alternative that enables the drone to reach its goal position or at least land safely in a pre-established location. New commercial implementations such as autonomous drone delivery could use the new localization method to improve the reliability of navigation.

The core attribute of the project is the environment in which vision-based localization is provided: \textit{in the wild}, denoting natural (non-urban) environments where artificial structures such as buildings and roads are sparse. This characteristic transforms what would be a trivial feature matching and homography computation problem inside a city into a challenging process, due to the difficulty of finding salient features in natural environments. Nevertheless, the final implementation is able to provide accurate localization results using a neural network for feature matching between drone photographs and satellite images.

Another significant feature of the implemented localization algorithm is that the pre-mapping process does not require the UAV to fly. The map is built using exclusively open-source satellite images, with the objective of enabling autonomous localization in any area where the drone can fly, assuming it is legally and physically possible (due to harsh environmental conditions).


\section{Methodology}
\label{sec:methodology}

\begin{figure}
    \centering 
    \captionsetup{justification=centering}
    \includegraphics[width=0.49\textwidth]{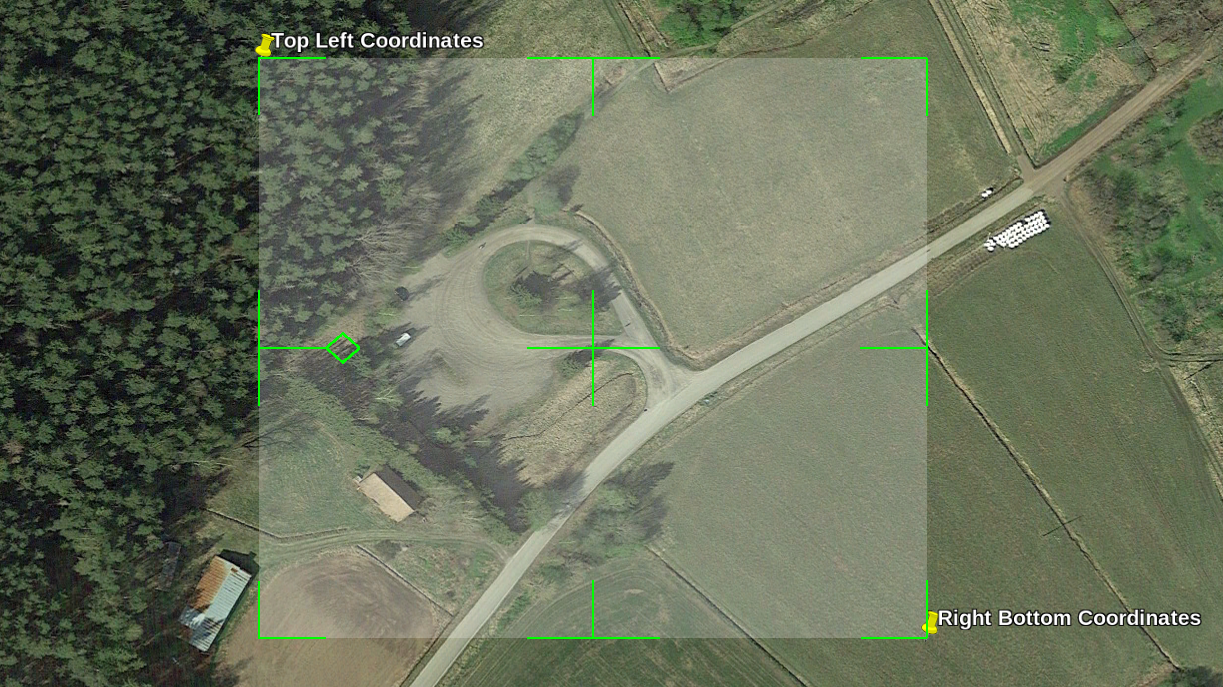}
    \caption{An example of a georeferenced section of the map used for image matching}
    \label{fig:map_patch} 
\end{figure}

Accurate feature matching of images is only useful if the drone camera photographs can be precisely linked to geographical coordinates. An important assumption of our proposed localization method is that the flight area of the UAV is known \textit{a priori}, so a map for that specific zone can be built and uploaded to the onboard computer for offline use. The map is composed of rectangular sections representing RGB satellite images with an approximate resolution of 1400×1200 pixels, collected from GE. Each one of these sections are collected from the same perspective, with the camera view perpendicular to the ground surface and from an altitude that offers a similar field of view to a wide angle camera. One example of these sections can be observed in Fig. \ref{fig:map_patch}, where the transparent white rectangle represents the georeferenced map tile. There is a linear relation (see Equations \ref{equation:latitude} and \ref{equation:longitude}) between the pixel coordinates of the image file and absolute geographical coordinates. 

\subsection{Georeferencing map sections}
\label{sec:georefenced_map}

Because the flight area is generally too large to be represented as only one image file, the map is split into different section, each one of them with two distinct geographical coordinates, a \textit{(latitude, longitude)} pair for the top left corner and another one for the right bottom corner. Instead of using a unique pair of coordinates that correspond to the center of the image, it is more accurate to geographically reference two corners of the section. In this manner, any set of features located in the image can be accurately linked to geographical coordinates with a simple equation that relates pixels to latitude:

\begin{equation}
  \mathbf{Lat} = Lat_t + \frac{C_y}{H}  * (Lat_b - Lat_t) 
  \label{equation:latitude}
\end{equation}

where \newline

\begin{xtabular}{ll}
    $\mathbf{Lat}$                &= computed latitude \\
    $Lat_t$                &= top left corner latitude \\
    $C_y$                &= vertical pixel coordinate of matched features  \\
    $H$                &= height of the satellite image in pixels \\
    $Lat_b$                &= right bottom corner latitude \\
    \newline
\end{xtabular}

Using the same logic, longitude is computed as follows:

\begin{equation}
  \mathbf{Lon} = Lon_t + \frac{C_x}{W}  * (Lon_b - Lon_t)  
  \label{equation:longitude}
\end{equation}

where \newline

\begin{xtabular}{ll}
    $\mathbf{Lon}$                &= computed longitude \\
    $Lon_t$                &= top left corner longitude \\
    $C_x$                &= horizontal pixel coordinate of matched features  \\
    $W$                &= width of the satellite image in pixels \\
    $Lon_b$                &= right bottom corner longitude \\
    $ $
\end{xtabular}

\begin{figure}
    \centering 
    \captionsetup{justification=centering}
    \includegraphics[width=0.49\textwidth]{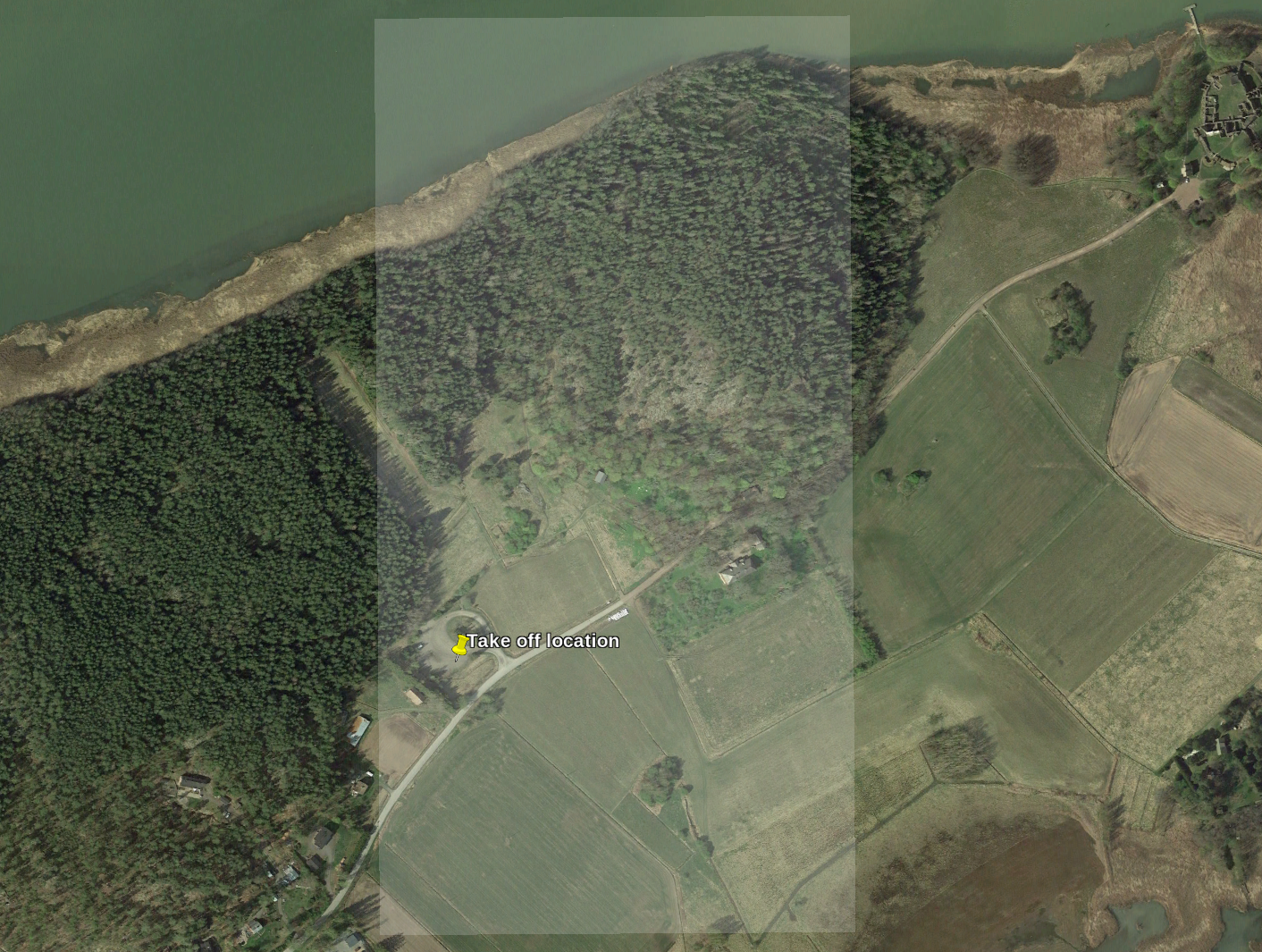}
    \caption{Satellite view of the flight zone (highlighted rectangle). The yellow pin is located at  60.403091$^{\circ}$ latitude and 22.461824$^{\circ}$  longitude }
    \label{fig:experiments_flight_area} 
\end{figure}

Since any set of matched features can be enclosed in a polygon in the georeferenced satellite image, the center of that polygon – represented by the $C_x$ and $C_y$ pixel coordinates – can be used to directly compute the geographical coordinates. In other words, the coordinate system is changed from pixels in a local image file to latitude and longitude.  This approach enables the algorithm to pinpoint accurately the geographical coordinates of an image region rich in salient features (e.g., a special shape such as the road in Fig. \ref{fig:experiments_flight_area}). Fig. \ref{fig:good_matches} shows examples of successfully matched drone-satellite image pairs, in spite of the relative scarcity of available features in the natural environment. Algorithm \ref{alg:localization_algorithm} presents the detailed sequence of computations needed for computing the geographical position of the UAV.

Fig. \ref{fig:experiments_flight_area} provides a bird's-eye view of the flight area chosen for gathering the drone photographs used for testing the developed algorithms. The zone was selected for a number of reasons, in particular because it contains a multitude of different surfaces (forests, fields, roads, sea shores, buildings) in a relatively small perimeter. In addition, it is also a sparsely populated area, which makes it suitable for drone experiments in accordance with EU regulation on the utilization of UAVs.

\subsection{Computing geographical coordinates}

\begin{algorithm}[t]
	\caption{Vision-based GNSS-free localization}
	\small
	\label{alg:localization_algorithm}
	\KwIn{ \\
	    \hspace{1em} RGB camera drone photograph: $dronePhoto$ \\
	    \hspace{1em} Pre-built satellite map with $N$ georeferenced photographs \\
	}
	\KwOut{\\
	    \hspace{1em} Absolute coordinates: $(latitude, longitude)$ \\
	}
	rotate$\left(dronePhoto, degrees = droneYaw\right)$
	
	\For {$satellitePhoto=1,N$}{
	    compute\_features$\left(dronePhoto, satellitePhoto\right)$
	    
	    matchesList =  match\_features$\left(dronePhoto, satellitePhoto\right)$
	    
	    \If{$length(matchesList) >= maxMatches$}{
	       maxMatches = $length(matchesList)$
	       
	       matchedPhoto = satellitePhoto
	       
	       transMat = findHomography\textit{(matchesList, \\ \hspace{1em}method = \scriptsize{RANSAC})}
	       
	       polygon = perspectiveTransform$\left(transMat\right)$
	       
	       pixelCoordinates = center\_of$\left(polygon\right)$
	    }
	}
	compute\_geo\_coordinates$\left(pixelCoordinates, matchedPhoto\right)$

\end{algorithm}

Algorithm \ref{alg:localization_algorithm} provides the core functionality of this project's software implementation. The developed method accepts as input RGB photographs (either JPEG or PNG) taken by the downward-facing drone camera and matches them against a pre-built map consisting of satellite georeferenced images (the map can be varying in size, depending on the size of the flying zone). Upon successful localization of a drone captured photograph, the output is a pair of absolute geographical coordinates (latitude and longitude), indicating the position of the aerial vehicle.  

In order to compute geographical coordinates, the algorithm uses a fairly simple approach. It starts by rotating the drone photograph using the image metadata providing the orientation of the UAV and the camera gimbal to match the heading of the map sections (always oriented northwards). The rotation is necessary because it improves the number of features which can be matched for a pair of images. Secondly, the features of both images (drone and satellite) are computed and matched against each other. This process is repeated for all the satellite images and the drone is assumed to be located in the map section which has the most matches with the current camera photograph. Following the feature matching process, a perspective transformation is computed, that provides a link between the position of the matched features in the georeferenced map image and their location in the drone photograph. Since the  matched map section is a rectangle with \textit{a priori} known geographical coordinates, it is possible to convert from pixel coordinates to latitude and longitude in a straightforward manner (see Equations \ref{equation:latitude} and \ref{equation:longitude} for the detailed calculation method).

An important parameter of the algorithm is the method used for differentiating between the matched features which represent inliers and the ones which are outliers. The method is called random sample consensus (RANSAC) and it improves accuracy of the localization algorithm by selecting only features which are relatively close together. These valid features are further used for the computation of the perspective transform and outliers are discarded.
The minimum number of matched features between a pair of photographs that enable the computation of the drone position is four, otherwise the perspective transform cannot be calculated. If the drone flies over featureless surfaces such as fields or forests, an incorrect section of the map could be matched with the camera photograph, resulting in erroneous geographical coordinates.



\section{Experimental Results}
\label{sec:results}

This section presents the experimental results for a photograph dataset on which the localization algorithm was tested. The photographs were taken on sunny weather in the wild (non-urban environment), from an altitude of approximately 120 meters AGL. Each photograph was captured from the same perspective (camera facing the ground surface). The camera used for gathering the dataset was a H20T, with a 12 Megapixel RGB sensor that has a 82.9$^{\circ}$ field of view and 4.5 mm focal length. The utilized UAV was a DJI Matrice 300, manually controlled by a human pilot inside the perimeter shown in Fig. \ref{fig:experiments_flight_area} The ground truth is provided by the GNSS metadata embedded into each photograph. 

\begin{figure}
    \centering
    \includegraphics[width=0.49\textwidth]{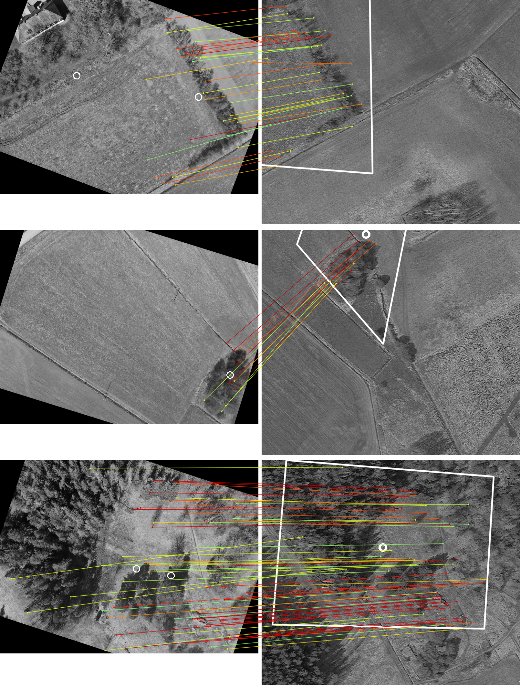} \\
    \small{\hfill(a) Query image\hfill\hspace{.8em}\hfill(b) Database image\hfill}
    \captionsetup{justification=centering}
    \caption{Examples of successfully matched drones photographs (a) and satellite images (b)}
    \label{fig:good_matches}
\end{figure}

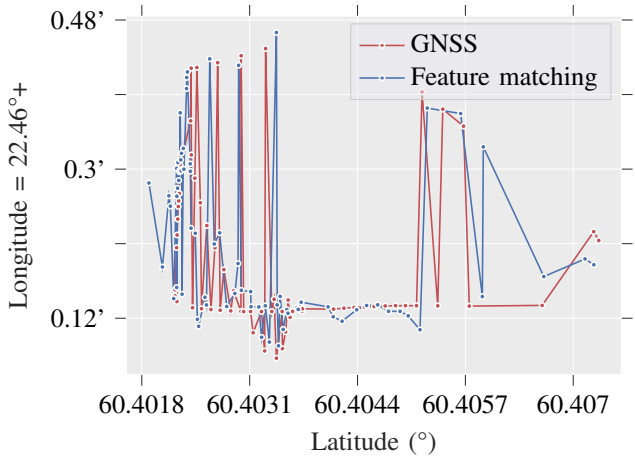
\begin{figure}
    \centering
    \setlength{\figurewidth}{.45\textwidth}
    \setlength{\figureheight}{.35\textwidth}
\begin{tikzpicture}

\definecolor{darkslategray38}{RGB}{38,38,38}
\definecolor{indianred1967882}{RGB}{196,78,82}
\definecolor{lavender234234242}{RGB}{234,234,242}
\definecolor{lightgray}{RGB}{235,235,235}
\definecolor{lightgray204}{RGB}{204,204,204}
\definecolor{steelblue76114176}{RGB}{76,114,176}

\begin{axis}[
height=\figureheight,
width=\figurewidth,
axis background/.style={fill=lightgray},
axis line style={white},
legend cell align={left},
legend style={
  fill opacity=0.5,
  draw opacity=1,
  text opacity=1,
  draw=lightgray204,
  fill=lavender234234242
},
tick align=outside,
x grid style={white},
xlabel=\textcolor{darkslategray38}{Latitude (\textdegree)},
xmajorgrids,
xmajorticks=true,
xmin=60.4016142617816, xmax=60.4075794795977,
xtick style={color=darkslategray38},
xtick={60.4018, 60.4031, 60.4044, 60.4057, 60.407},
xticklabels={60.4018, 60.4031, 60.4044, 60.4057, 60.407},
y grid style={white},
ylabel=\textcolor{darkslategray38}{Longitude = 22.46\textdegree+},
ymajorgrids,
ymajorticks=true,
ymin=22.4608696392083, ymax=22.4680764655139,
ytick style={color=darkslategray38},
ytick={22.462,  22.4635, 22.465,  22.4665,  22.468 },
yticklabels={0.12', , 0.3', ,   0.48' }
]
\path [draw=indianred1967882, fill=indianred1967882, opacity=0.2]
(axis cs:60.4022555555556,22.4650555555556)
--(axis cs:60.4022555555556,22.4648333333333)
--(axis cs:60.4022583333333,22.4650484567901)
--(axis cs:60.4022611111111,22.4650527777778)
--(axis cs:60.4023888888889,22.46575)
--(axis cs:60.4023888888889,22.4661916666667)
--(axis cs:60.4023888888889,22.4661916666667)
--(axis cs:60.4022611111111,22.4651631944444)
--(axis cs:60.4022583333333,22.4651234722222)
--(axis cs:60.4022555555556,22.4650555555556)
--cycle;

\path [draw=steelblue76114176, fill=steelblue76114176, opacity=0.2]
(axis cs:60.4022357255747,22.46515949)
--(axis cs:60.4022357255747,22.46510522)
--(axis cs:60.4022446508621,22.4645308625)
--(axis cs:60.4022558074713,22.4647298525)
--(axis cs:60.4022558074713,22.46515949)
--(axis cs:60.4022558074713,22.46515949)
--(axis cs:60.4022446508621,22.46501477)
--(axis cs:60.4022357255747,22.46515949)
--cycle;
\path [draw=steelblue76114176, fill=steelblue76114176, opacity=0.2]
(axis cs:60.4022781206897,22.4654172725)
--(axis cs:60.4022781206897,22.465222805)
--(axis cs:60.4022781206897,22.4654172725)
--(axis cs:60.4022781206897,22.4654172725)
--cycle;
\path [draw=steelblue76114176, fill=steelblue76114176, opacity=0.2]
(axis cs:60.4023071278736,22.4650735625)
--(axis cs:60.4023071278736,22.46492432)
--(axis cs:60.4023071278736,22.4650735625)
--(axis cs:60.4023071278736,22.4650735625)
--cycle;

\addplot [semithick, indianred1967882, mark=*, mark size=1, mark options={solid,draw=white}]
table {%
60.4021972222222 22.46245
60.4022 22.4625722222222
60.4022194444444 22.4634194444444
60.4022222222222 22.4623361111111
60.402225 22.463675
60.4022333333333 22.4639944444444
60.4022388888889 22.4642472222222
60.4022444444444 22.4643611111111
60.4022527777778 22.4645083333333
60.4022555555556 22.4649444444444
60.4022583333333 22.4650805555556
60.4022611111111 22.4650902777778
60.4023888888889 22.4659708333333
60.4023944444444 22.4670305555556
60.4023972222222 22.4652861111111
60.4024111111111 22.4622111111111
60.4024388888889 22.4648166666667
60.4024638888889 22.4670444444444
60.4025055555556 22.4643222222222
60.4025194444445 22.4621916666667
60.4025833333333 22.4638638888889
60.4026333333333 22.462175
60.4026805555556 22.4634166666667
60.4027138888889 22.4671416666667
60.4027444444445 22.4621611111111
60.4027888888889 22.4629777777778
60.4028722222222 22.4621472222222
60.4029055555556 22.46255
60.4029944444444 22.4621388888889
60.4029972222222 22.4672805555556
60.4030277777778 22.4621333333333
60.4031111111111 22.4621361111111
60.4031416666667 22.4617083333333
60.4032416666667 22.4621333333333
60.4032805555556 22.4613416666667
60.4032916666667 22.467425
60.4033666666667 22.4621361111111
60.4033944444445 22.4623805555556
60.4034222222222 22.4611972222222
60.4034916666667 22.4621361111111
60.4034944444444 22.4613888888889
60.4035333333333 22.461725
60.4035638888889 22.4623666666667
60.4035861111111 22.4620194444444
60.4036166666667 22.4621361111111
60.4037333333333 22.4621444444444
60.4037361111111 22.4621888888889
60.4041111111111 22.4621833333333
60.4042361111111 22.4622027777778
60.4043583333333 22.4622166666667
60.4044833333333 22.4622277777778
60.4046 22.4622361111111
60.4047333333333 22.4622472222222
60.4048583333333 22.4622527777778
60.4049777777778 22.4622527777778
60.4051111111111 22.4622527777778
60.4051777777778 22.4665527777778
60.4053666666667 22.46225
60.4054277777778 22.4662055555556
60.4056777777778 22.4658638888889
60.4057472222222 22.4622444444444
60.406625 22.4622583333333
60.4072472222222 22.4637416666667
60.4073083333333 22.4635638888889
};
\addlegendentry{GNSS}
\addplot [semithick, steelblue76114176, mark=*, mark size=1, mark options={solid,draw=white}]
table {%
60.401885408046 22.4647208075
60.4020485258621 22.4630313925
60.4021263908046 22.464463025
60.4021442413793 22.46425499
60.4021843427506 22.4623956325697
60.4022201063218 22.46501477
60.4022203376437 22.462619845
60.4022223376437 22.46445398
60.402231262931 22.46461679
60.4022357255747 22.465132355
60.4022446508621 22.46477281625
60.4022558074713 22.46497256
60.4022629006659 22.4661325919222
60.4022691954023 22.465186625
60.402273658046 22.4649650225
60.4022781206897 22.46532003875
60.402285045977 22.46248417
60.402300433908 22.4654172725
60.4023071278736 22.46499894125
60.4023394616943 22.4666200487814
60.4023413268805 22.4668433013777
60.4023495229885 22.4669549225
60.402383617811 22.4651055691667
60.4023852241379 22.4649559775
60.4023925520498 22.4638116873924
60.4024440155364 22.4637096388094
60.4024694193084 22.46197865
60.4024850835735 22.4618384525
60.402561167147 22.462421855
60.4025790691643 22.46226809
60.4026185442183 22.4672183083333
60.402673054755 22.4634936875
60.4027379463087 22.4637184042438
60.4028229841499 22.4622364325
60.402921445245 22.4624987375
60.4029594870317 22.46310023
60.4029666915203 22.4670880928573
60.4029997665706 22.4625620525
60.4031116541787 22.4625349175
60.403116129683 22.46223191
60.4032101152738 22.4622273875
60.4032436815562 22.46161685
60.4032884365994 22.46225
60.403337667147 22.4615218775
60.4034223544669 22.4677488825
60.4034316527378 22.46226809
60.4034473170029 22.461444995
60.4034674567723 22.462439945
60.4035032608069 22.4617751375
60.4035480158501 22.4621731175
60.4035614423631 22.4621007575
60.4036666167147 22.4622092975
60.403684518732 22.46217764
60.4037203227666 22.46232236
60.4040443753582 22.4622293875
60.4041067134671 22.4620303975
60.4042135787966 22.4619399475
60.404389461318 22.462170595
60.4045096848138 22.4622565225
60.4046432664756 22.46227009
60.4047746217765 22.4621389375
60.4049126561605 22.4621389375
60.4050106160458 22.4620484875
60.4051531031519 22.4617680925
60.4052409690577 22.4662283225
60.4056428612303 22.466117215
60.4059043561254 22.4624359
60.4059161359084 22.465447885
60.406645460114 22.4628384025
60.4071426467176 22.4631932959375
60.4072486394357 22.4630755963365
};
\addlegendentry{Feature matching}
\end{axis}

\end{tikzpicture}
    \caption{Computed coordinates and GNSS ground truth }
    \label{fig:abs_coord}
\end{figure}

\begin{figure}
    \centering
    \setlength{\figurewidth}{.42\textwidth}
    \setlength{\figureheight}{.35\textwidth}
\begin{tikzpicture}

\definecolor{darkslategray38}{RGB}{38,38,38}
\definecolor{indianred1967882}{RGB}{196,78,82}
\definecolor{lightgray}{RGB}{235,235,235}

\begin{axis}[
height=\figureheight,
width=\figurewidth,
axis background/.style={fill=lightgray},
axis line style={white},
tick align=outside,
x grid style={white},
xlabel=\textcolor{darkslategray38}{Photograph Index},
xmajorgrids,
xmajorticks=true,
xmin=-6.15, xmax=129.15,
xtick style={color=darkslategray38},
y grid style={white},
ylabel=\textcolor{darkslategray38}{Localization Error [m]},
ymajorgrids,
ymajorticks=true,
ymin=-1.99263327108, ymax=50.4833199186772,
ytick style={color=darkslategray38}
]
\addplot [semithick, indianred1967882, mark=*, mark size=1, mark options={solid,draw=white}]
table {%
0 2.2684709869351
1 12.1337798031663
3 2.6542667544331
6 45.5814182647215
8 5.16411363021586
9 4.73553359007691
10 20.4900446505478
11 5.86737170807926
12 8.90000487443478
13 7.44946612051353
14 3.82291970521122
15 5.54449072223856
16 2.5470839164014
17 0.392637328454413
18 21.1590240682796
19 12.1705186126763
20 9.32279454422899
21 20.3144176030829
22 17.4038547663442
23 44.6459199821524
24 28.5552103230092
30 3.45756031413978
31 3.31156017577463
32 10.720061441008
35 22.0573640853464
36 12.9539230191469
37 7.90855673379133
38 24.9075308054058
39 7.34707979850406
40 21.3645585103764
41 21.9019937746002
42 6.24411856467072
43 10.7139584862417
44 9.85221708861095
45 7.89953444224241
46 8.22920822796084
49 7.83984953096053
50 17.2232907713907
51 22.1362702262577
52 10.9002467634152
53 10.1049669040875
54 10.0934188794471
55 11.2128293106124
56 9.56648124493711
57 15.8344369905962
59 35.5565958192587
62 20.3917809829519
69 31.9343122078629
78 27.6223134286863
79 32.2792459922898
87 34.9906560880633
88 24.403271122059
89 19.1529580231031
96 22.9685390332225
97 11.1016441884446
98 11.406842113472
99 13.6342389925154
100 11.8594265102747
103 14.3800447891238
104 48.0980493191428
105 10.0300732330399
106 9.70282675459225
107 30.7253184128705
108 17.6564185877845
109 17.7575331081662
110 31.1236357385126
111 30.8073692014986
112 23.7498777427817
113 28.8939811732276
114 15.6589668431922
115 20.1566402232034
116 6.07907789155449
117 4.33165854615196
118 5.24073816379562
119 5.76998437520518
122 17.5640365976027
123 8.74926313088702
};
\end{axis}

\end{tikzpicture}
    \caption{Absolute localization error}
    \label{fig:abs_loc_error}
\end{figure}
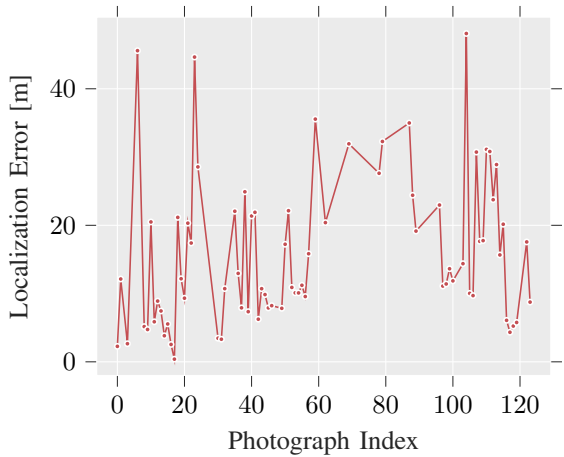

The overall performance of the vision-based localization algorithm can be summarized by  Fig. \ref{fig:abs_loc_error}. The error is calculated by considering the proximity of the computed geographical coordinates to the ground truth, represented by the GNSS metadata for each photograph. A comparison between the absolute GNSS coordinates (ground truth) and the coordinates provided by our proposed localization algorithm can be seen in Fig. \ref{fig:abs_coord}. The dataset consists of 126 photographs in total, from which 77 were successfully located using our vision-based algorithms.  A photograph was considered successfully localized if the error between the ground truth GNSS position of the drone and the location computed by the vision-based algorithm was less than 50 meters.  For the whole dataset, a mean average error (MAE) of 15.82 meters suggests the new localization method has an accuracy that can be compared to that of standard GNSS positioning methods, for which the error can range from 1 to 20 meters under clear sky conditions~\cite{wing2005consumer}, depending on other factors such as the available number of satellites. The coordinates determined by the vision-based localization algorithm closely follow the GNSS ones, with a few exceptions where the error exceeds 30 meters, as can be seen in Fig. \ref{fig:abs_loc_error}. Admittedly, the accuracy is not high enough to be used in precision maneuvers or at low altitudes where obstacles are present, but the algorithm represents an important building block towards GNSS-Free navigation strategies for UAVs.

Since the feature matching model is not rotation invariant, it was necessary to use the metadata of the drone photographs to align them with the map images, which have the same heading (north). Because the UAV compass suffered from a systematic error, all the recorded photographs were rotated with 15$^{\circ}$  clockwise in addition to the gimbal and drone yaw. This small correction reduced the misalignment between the drone photographs and their satellite counterparts, maximizing the performance of the feature matching algorithm.

The MAE value (15.82 m) is of  particular interest because it shows that most of the images for the first experiment are located with an error comparable to traditional GNSS services. The computed geographical coordinates feature significantly increased errors, which suggests that navigation strategies must include mechanisms that allow the UAV to fly only over surfaces that offer salient features to be matched. The performance of the vision-based localization algorithm is directly dependent on the trajectory computed by the path planner, as suggested in ~\cite{bartolomei2020perception}.

There are a number of important factors which affect the accuracy of the vision-based localization algorithm. Among these are weather and light – depending on the season and the amount of lighting, the drone photographs could have more or less features that can be matched with the onboard satellite map. The photos taken during the experiments were taken one year later, but in the same month (May) as the satellite images, in order to maximize the similarity between them. Moreover, flying above featureless surfaces such as empty fields or water can render the vision-based localization algorithm ineffective since it is based on recognition of salient features.

Another important aspect is that artificial structures such as buildings and roads tend to have more features that can be tracked and matched, compared to natural objects. This is the reason that transforms feature matching into a challenging task when performed in the wild (outside urban environments). In addition to this, the open source photographs for non-urban environments 
are characterized by lower resolution and are more rarely updated than data available for cities. Nonetheless, Fig. \ref{fig:good_matches} shows that the model developed in this project is able to provide accurate matching results even when the drone photographs differ in perspective, rotation and contain surfaces sparse in salient features.


\section{Conclusion}\label{sec:conclusion}

The goal of this paper was to demonstrate that GNSS-Free vision-based localization of UAVs flying in the wild was not only possible, but a viable method which provides accurate results. To achieve this objective, a robust localization algorithm was implemented. To prove its effectiveness, the vision-based localization algorithm was tested on an experimental dataset consisting of RGB photographs taken with a drone flying at high altitude in the wild. 

The most important feature to be researched in the future is adding the possibility of using a LIDAR to improve the localization capabilities of the algorithm in different lighting scenarios. This approach would enable localization of the UAV even during night or harsh weather conditions, significantly extending the applicability of the studied GNSS-Free localization algorithm.

Another improvement would be the implementation of a path planner that dictates the trajectory of the UAV, given an initial position and an end goal. The path planner would make use of the developed vision-based localization algorithm  and also devise a navigation strategy to avoid flying the drone above featureless areas, where pose estimation using visual techniques becomes challenging or impossible. 

An incremental improvement of the current implementation, given enough resources, could be done by retraining the graph neural network used in feature matching using RGB photographs taken with the drone camera in different non-urban environments. Although time-consuming, the retraining process would significantly improve the accuracy of the localization algorithm.


\section*{Acknowledgment}

This research work is supported by the Academy of Finland's AutoSOS and AeroPolis projects (Grant No. 328755 and 348480) and by the Finnish Ministry of Defence's Scientific Advisory Board for Defence (MATINE) project WildNav.

\bibliographystyle{unsrt}
\bibliography{bibliography}

\end{document}